\PassOptionsToPackage{dvipsnames}{xcolor}
\documentclass[10pt]{cai27}
\usepackage{graphicx}
\usepackage{booktabs}
\usepackage{longtable}
\usepackage{url}
\usepackage{fvextra}
\usepackage{tabularx}
\usepackage{placeins}
\usepackage{flafter}
\usepackage{xurl}
\usepackage{microtype}
\AtBeginBibliography{%
    \setlength{\emergencystretch}{3em}%
}

\begin{document}

\begin{center}
    \title{QFrBLiMa: a Quebec French Benchmark of Regional Linguistic Markers}
    \maketitle
    \thispagestyle{empty}
    \pagenumbering{arabic}


    \begin{tabular}{cc}
        Nasma Chaoui\upstairs{\affilone,*}, Richard Khoury\upstairs{\affilone, **} \\[0.25ex]
        {\small \upstairs{\affilone} Université Laval}
    \end{tabular}
    \emails{\upstairs{*} nasma.chaoui.1@ulaval.ca \\ \upstairs{**} richard.khoury@ift.ulaval.ca}
    
    \vspace*{0.1in}

\end{center}

\begin{abstract}
Recognizing regional linguistic variation is essential for developing language technologies that account for the diversity of language use. We introduce QFrBLiMa, a contrastive benchmark for evaluating the recognition of documented regional associations of linguistic markers in Quebec French and Metropolitan French. Grounded in the \textit{Grand dictionnaire terminologique}, it comprises 1,205 sentence pairs covering 241 linguistic markers across 12 thematic categories. We evaluate two widely used classification strategies: zero-shot prompting of large language models (LLMs) and supervised fine-tuning of BERT-family classifiers. Our experiments involve eight LLMs and five encoders fine-tuned using a separate corpus of 938,484 labeled text samples. Under their respective evaluation protocols, Gemini achieves the highest LLM order-robust accuracy (93.53\%), while CamemBERT achieves the highest encoder strict pairwise accuracy (72.45\%). The two strategies exhibit complementary recognition patterns. Our analyses show that stronger preferences do not necessarily accompany higher accuracy, candidate order substantially affects several LLMs, and source-based classification transfers imperfectly to regional lexical recognition. Moreover, identifying the relevant variants does not guarantee correct regional attribution. These findings motivate dedicated contrastive evaluation of regional linguistic knowledge beyond general multilingual capabilities. The methodology can be adapted to other linguistic varieties. The benchmark, code, and experimental results are publicly available on GitHub\footnote{\url{https://github.com/chaouin/QFrBLiMa}}, and the benchmark and five fine-tuned classifiers are available on Hugging Face\footnote{\url{https://huggingface.co/collections/chaouin/qfrblima}}.
\end{abstract}

\begin{keywords}{Keywords:} Quebec French, regional variation, contrastive evaluation, large language models, supervised classification.\end{keywords}

\section{Introduction}
Using the appropriate regional vocabulary matters for translation and writing assistance. For instance, the words \textit{puzzle} and \textit{casse-t\^ete} have the same meaning, but one is characteristic of Quebec French (QFr) and the other of Metropolitan French (MFr), and so the choice of which one to use depends on the intended public of the document. Grammatical acceptability and semantic understanding are distinct challenges from regional attribution.

In this paper, we present QFrBLiMa, a dataset comprising 241 linguistic markers, each consisting of two variants (words or expressions) with the same intended meaning and different documented regional associations. Each marker is used in five contexts, or sentence pairs differing only by their choice of variant and related grammatical adjustments and labelled by region.

We then introduce the task of regional sentence attribution, and benchmark example systems of two widely used natural language processing (NLP) classification strategies \citep{devlin-etal-2019-bert,edwards-camacho-collados-2024-language}: zero-shot LLM prompting and supervised BERT-family fine-tuning. We investigate (i) how accurately and with what strength of preference the models recognize regional associations, (ii) how consistently they recognize these associations across candidate orders (for LLMs) and contexts, (iii) whether source-based classification transfers to regional lexical recognition and how performance relates to reported pretraining characteristics, and (iv) which thematic and linguistic difficulties characterize recognition errors. Rather than isolating the effect of model architecture or supervision, we compare the two strategies as distinct practical approaches to regional linguistic recognition.

Our contributions are a source-grounded benchmark of 1,205 sentence pairs; an evaluation of 13 models comprising 19,280 prompted judgments and 6,025 encoder pair rankings; and an analysis of complementary coverage, preference strength, candidate-order sensitivity, contextual consistency, classification transfer, thematic recognition and regional attribution errors. Exhaustive candidate reversal distinguishes robust recognition from success under a single presentation order, for which balanced answer positions otherwise allow a fixed-position predictor to approach 50\% accuracy \citep{liusie-etal-2024-llm,pezeshkpour-hruschka-2024-large}. Together, these analyses characterize the strengths and limitations of the two strategies beyond aggregate accuracy.


\section{Related Work}
\label{sec:relatedwork}

\subsection{Contrastive Evaluation and Quebec French Benchmarks}

Contrastive benchmarks such as BLiMP and MultiBLiMP evaluate the performance of NLP tools on specific linguistic phenomena using minimal sentence pairs \citep{warstadt2020blimp,jumelet-etal-2026-multiblimp}. These benchmarks have been implemented in a variety of languages and dialects, including Quebec French, where QFrBLiMP assesses grammatical acceptability \citep{beauchemin-etal-2026-qfrblimp} using documented phenomena taken from the \textit{Office québécois de la langue française (OQLF)}.

\subsection{Regional Variation and Dialect Identification}

In parallel, multiple studies, such as DialectBench, have highlighted performance disparities across linguistic varieties and motivate evaluation beyond dominant varieties \citep{faisal-etal-2024-dialectbench}. In French, this has led to the creation of resources such as Carte, which evaluates LLM knowledge of regional differences within France \citep{carneiro2026carte}, and FreCDo, which distinguishes Belgian, Canadian, Metropolitan and Swiss French across domains \citep{gaman2023frecdo}.

Benchmarks of Quebec and Metropolitan French idioms expand this distinction to cultural elements \citep{beauchemin-etal-2026-idiom}.

QFrBLiMa complements these benchmarks by evaluating whether models can distinguish the documented regional associations of lexical and terminological variants intended to express the same meaning. Its contrastive format therefore targets a capability distinct from both grammatical acceptability and semantic understanding.


\section{QFrBLiMa Dataset}

We expand the literature of Section \ref{sec:relatedwork} by introducing the challenge of regional attribution, or of determining which dialectal region a sentence originates from. This task is vitally important when adapting LLMs or other NLP tools to serve the people of a specific region. Indeed, two sentences can be equally grammatical and carry the same meaning, but one can seem ``usual'' and the other ``odd'' because they use linguistic markers that the users are familiar or unfamiliar with. Table \ref{tab:QFrBLiMa example} gives two examples of this phenomenon. These sentences differ only by their choice of linguistic markers (\textit{puzzle}/\textit{casse-tête}, \textit{vacances universitaires}/\textit{relâche}) and the resulting grammatical modifications (\textit{vacances universitaires} is feminine plural while \textit{relâche} is feminine singular), but each one would sound ``out of place'' outside the geographic region each marker originates from.

\begin{table}[!htbp]
\centering
\begin{tabular}{l|l}
\textbf{Sentence} & \textbf{Label} \\
\hline
\textit{J'ai terminé le puzzle de mille pièces hier soir.} & MFr \\
\textit{J'ai terminé le casse-tête de mille pièces hier soir.} & QFr \\
\hline
\textit{Les vacances universitaires commencent lundi.} & MFr \\
\textit{La relâche commence lundi.} & QFr \\
\hline
\end{tabular}
\caption{Sample minimal pairs from QFrBLiMa.}
\label{tab:QFrBLiMa example}
\end{table}

We document this phenomenon by creating QFrBLiMa, the \textbf{Q}uebec \textbf{Fr}ench \textbf{B}enchmark of \textbf{Li}nguistic \textbf{Ma}rkers.

\subsection{Source and Linguistic Marker Extraction}

We use the January 2026 snapshots of two datasets from the \textit{Grand dictionnaire terminologique} (GDT), which contains 53,899 terminological records and a supplementary dataset of officially recognized terms documented by experts from the OQLF and external partners under established editorial guidelines \citep{oqlf_data,oqlf_policy2022}. Its documented meanings and geographic usage provide the reference associations for QFrBLiMa.

We use a Python script to extract individual terminological variants from the French-language field of each record, remove grammatical annotations, normalize terms to identify duplicates, and retain geographic indications, definitions, domains and source URLs. Within each entry, we pair distinct terms associated with Quebec/Canada (which we label as QFr in our dataset) or France/Europe (labelled as MFr). Terms without explicit geographic indications are excluded. This extracts 253 linguistic markers of labelled word pairs.

Manual review reduces the set to 241 linguistic markers, primarily to eliminate redundancy arising from masculine and feminine variants of the same terms. Only the masculine forms are retained when such variants coexist. All retained linguistic markers have a source URL. Note that the QFr and MFr labels reflect documented geographic associations rather than geographically exclusive usage.

\subsection{Sentence Pair Construction and Thematic Categories}

For each linguistic marker, we manually write five sentence contexts based on the meaning documented in the corresponding GDT definition, rather than extracting sentences directly from the GDT. We construct each sentence pair by substituting the regional term and adjusting agreement, determiners or elision when necessary, as shown in Table \ref{tab:QFrBLiMa example}. Using multiple contexts allows us to distinguish recognition that is consistent across different sentence realizations from recognition that depends on a particular context. We did not perform independent validation of contexts for naturalness, semantic equivalence or sense disambiguation. The 1,205 resulting pairs average 9.01 MFr and 8.96 QFr whitespace-separated words.

The linguistic markers originate from 157 GDT entries with 125 distinct domain-label strings. We retain the first top-level domain (51 principal domains) and manually aggregate these domains into 12 mutually exclusive thematic categories based on semantic proximity to reduce fragmentation. For example, \textit{sport}, \textit{loisir}, and \textit{plein air} are grouped under \textit{Sports and Recreation}, while \textit{botanique}, \textit{zoologie}, and \textit{agriculture} are grouped under \textit{Science and Nature}. The complete domain-to-category mapping is provided in the accompanying repository.

\section{Experimental Setup}
\subsection{Evaluated Models}
\label{sec:encoders_method}
We evaluate three commercial LLMs (Gemini 3.8 Flash, Claude Opus 4.8, GPT-5.4) and five open-weight models (Llama 3.1 8B Instruct\footnote{\url{https://huggingface.co/meta-llama/Llama-3.1-8B-Instruct}}, Aya Expanse 8B\footnote{\url{https://huggingface.co/CohereLabs/aya-expanse-8b}}, Qwen3 8B\footnote{\url{https://huggingface.co/Qwen/Qwen3-8B}}, Mistral NeMo 12B Instruct\footnote{\url{https://huggingface.co/mistralai/Mistral-Nemo-Instruct-2407}}, Gemma 4 12B IT\footnote{\url{https://huggingface.co/google/gemma-4-12B-it}}), covering different families, multilingual capabilities and accessibility. We also evaluate five BERT-style encoders:  CamemBERT\footnote{\url{https://huggingface.co/almanach/camembert-base}}, CamemBERTv2 (RoBERTa-based)\footnote{\url{https://huggingface.co/almanach/camembertv2-base}}, ModernCamemBERT\footnote{\url{https://huggingface.co/almanach/moderncamembert-base}}, BERT-base-cased\footnote{\url{https://huggingface.co/google-bert/bert-base-cased}} and RoBERTa-base\footnote{\url{https://huggingface.co/FacebookAI/roberta-base}}. All encoders use a binary sequence classification head; the three French-pretrained and two English-pretrained models receive identical downstream supervision. Differences in architecture, tokenization and pretraining prevent an isolated comparison of pretraining language. The accompanying repository documents the API model identifiers, inference configurations and pretrained checkpoint references.

\textbf{Pretraining exposure.} Table~\ref{tab:pretraining} summarizes the available pretraining statistics for the evaluated models. Llama 3.1 and Qwen3 report large multilingual pretraining budgets \citep{grattafiori2024llama, yang2025qwen3}, while French-specific encoders provide more detailed information on their French training data \citep{martin2020camembert,antoun2024camembert,antoun2025modern}. Comparable French-specific statistics could not be established for the evaluated LLMs. All five encoders additionally receive 751,288 labeled French examples, including 375,644 examples per regional variety, whereas the LLMs receive no task-specific fine-tuning in our experiments. These figures contextualize the comparison but do not establish differences in actual French or Quebec French exposure across strategies.

\begin{table}[!htbp]
\centering\small

\begin{tabularx}{\textwidth}{p{0.30\textwidth}p{0.25\textwidth}X}
\toprule
Model & Reported pretraining scale & French pretraining exposure \\
\midrule
\multicolumn{3}{l}{\textit{Strategy 1: Zero-shot LLMs}} \\
Llama 3.1 8B
    & $>$15T tokens
    & Multilingual; French share undisclosed \\
Qwen3 8B
    & $\approx$36T tokens
    & 119 languages and dialects; French share undisclosed \\
Mistral NeMo 12B
    & Undisclosed
    & Multilingual; French share undisclosed \\
Aya Expanse 8B
    & Undisclosed
    & 23 target languages, including French \\
Gemma 4 12B IT
    & Undisclosed
    & 140+ pretraining languages; French share undisclosed \\
Gemini 3.8 Flash, Claude Opus 4.8, GPT-5.4
    & Not established
    & Not quantified \\
\midrule
\multicolumn{3}{l}{\textit{Strategy 2: Fine-tuned encoders}} \\
CamemBERT
    & 138 GB ($\approx$32.7B subword tokens)
    & French corpus \\
CamemBERTv2
    & 275B-token corpus
    & French corpus \\
ModernCamemBERT 
    & 1T-token dataset 
    & French corpus \\
BERT-base-cased
    & $\approx$3.3B words
    & English corpus \\
RoBERTa-base
    & $\approx$160 GB
    & English corpus \\
\bottomrule
\end{tabularx}
\caption{Reported pretraining scale and French-language exposure.
B and T denote billions and trillions, respectively.
LLM token counts refer to model-family disclosures.
Corpus sizes, word counts, and processed-token budgets are
not directly comparable. Language coverage does not establish
language-specific training proportions. Quebec-French-specific
exposure remains unquantified for all evaluated models.}
\label{tab:pretraining}
\end{table}

\subsection{Supervised Corpus and Fine-Tuning}

The five BERT models are fine-tuned to Quebec/Metropolitan dialect recognition. Previous studies demonstrate the effectiveness of CamemBERT fine-tuning for French dialect identification \citep{bernier-colborne-etal-2022-transfer}, while acknowledging that texts may be compatible with multiple regional varieties \citep{bernier-colborne-etal-2023-dialect}. These studies inform our supervised classification approach.

We construct a regional classification corpus covering six genres to expose the BERT classifiers to diverse linguistic registers and communicative contexts. Our approach is informed by the cross-domain dialect identification framework of FreCDo \citep{gaman2023frecdo}. Following the data source selection of \citet{khan2026quebec}, we build a Quebec-French corpus using geographically selected Wikipedia articles, news articles from CN2i and Radio-Canada, oral transcripts from CRIFUQ, and informal online discussions from Quebec Reddit communities and Depotoir.ca. We also collect parliamentary proceedings from the National Assembly of Quebec\footnote{\url{https://www.bibliotheque.assnat.qc.ca/fr/1793-journal-des-debats-1867-a-ce-jour}}, as well as administrative reports from the Auditor General of Quebec\footnote{\url{https://www.vgq.qc.ca/fr/publications}}. For Metropolitan French, we select sources covering the same genres: geographically selected Wikipedia articles excluding Quebec-related articles, oral transcripts from ORFEO \citep{orfeo}, computer-mediated communication from CoMeRe \citep{comere}, French news articles from MLSUM \citep{scialom-etal-2020-mlsum}, parliamentary proceedings from the French National Assembly\footnote{\url{https://www.assemblee-nationale.fr/dyn/17/comptes-rendus/seance}} and administrative reports from Vie-publique\footnote{\url{https://www.vie-publique.fr/rapports}}.

The resulting corpus contains 938,484 labeled text samples, partitioned into training, validation, and test sets (Table~\ref{tab:corpus_sources}). Labels are assigned at the source level rather than through individual linguistic marker annotation. Genre sizes reflect data availability, with smaller parliamentary and oral subsets. We ensure equal representation of MFr and QFr within each genre and partition rather than enforcing uniformity across genres. Document grouping is applied where identifiers exist, and row identifiers are used otherwise. Matching genres reduces genre-related confounds but does not eliminate source-specific effects. QFrBLiMa is constructed separately and excluded from training and checkpoint selection; upstream overlap cannot be ruled out.

Permission was obtained to use Quebec French resources that were not freely available for reuse. Following \citet{khan2026quebec}, we do not redistribute the supervised corpus or its partitions. Detailed source documentation, access conditions, aggregate statistics, and corpus preparation scripts are provided in the accompanying repository.

\begin{table}[!htbp]
\centering\footnotesize
\setlength{\tabcolsep}{4pt}
\begin{tabularx}{\textwidth}{lXXrrr}
\toprule
& & & \multicolumn{3}{c}{Text samples per variety} \\
\cmidrule(lr){4-6}
Genre & MFr sources & QFr sources & Train & Val. & Test \\
\midrule
Encyclopedic
    & Wikipedia (French category, QC excl.)
    & Wikipedia (QC category)
    & 80,000 & 9,999 & 10,000 \\
Parliamentary
    & French National Assembly
    & National Assembly of Quebec
    & 28,000 & 3,500 & 3,500 \\
Administrative reports
    & Vie-publique
    & Auditor General of Quebec
    & 79,999 & 10,000 & 10,000 \\
Oral
    & ORFEO
    & CRIFUQ
    & 27,808 & 3,386 & 3,429 \\
Informal web
    & CoMeRe
    & Reddit (QC communities), Depotoir.ca
    & 79,837 & 9,999 & 9,785 \\
Journalistic
    & MLSUM
    & CN2i, Radio-Canada
    & 80,000 & 10,000 & 10,000 \\
\midrule
\multicolumn{3}{l}{Total per variety}
    & 375,644 & 46,884 & 46,714 \\
\multicolumn{3}{l}{\textbf{Total (both varieties)}}
    & \textbf{751,288} & \textbf{93,768} & \textbf{93,428} \\
\bottomrule
\end{tabularx}
\caption{Sources and distribution of the supervised regional classification corpus. Sample counts are identical for MFr and QFr within each genre and partition. The corpus contains 938,484 labeled text samples.}
\label{tab:corpus_sources}
\end{table}

Both encoder and head parameters are optimized using two-class cross-entropy, $-\log\mathrm{softmax}(z(x))_y$, where $y=0$ denotes MFr and $y=1$ denotes QFr. All models use three epochs, seed 42, AdamW, learning rate $2\times10^{-5}$, linear scheduling with 0.06 warmup ratio, effective batch size 32 (16 with two accumulation steps), maximum length 128 tokens and weight decay 0.01. Validation macro-F1 selects epoch 2 for ModernCamemBERT and epoch 3 for the other models. Training optimizes individual classification, not contrastive ranking. All five fine-tuned classifiers are publicly available through the QFrBLiMa collection on Hugging Face.

\subsection{Evaluation Protocols}

\textbf{Zero-shot LLMs.} Pairwise judgments can be sensitive to candidate placement and answer-label preferences \citep{liusie-etal-2024-llm,pezeshkpour-hruschka-2024-large,wang-etal-2024-large-language-models-fair}. This is particularly relevant to contrastive benchmarks, where changing the presentation order should not alter the underlying linguistic distinction. Each prompt presents a QFrBLiMa sentence pair in randomized order as \texttt{A} and \texttt{B} and asks the model to pick the Quebec French one. The model must choose between five options, either a strong or slight preference for \texttt{A} or \texttt{B}, or \texttt{EQUIVALENT}. LLMs return a JSON with their choice, the dominant linguistic dimension, evidence spans and rationale (empty spans for equivalence). Instructions cover colloquial, standard, institutional, technical and formal registers and ask models to disregard candidate order and avoid invented differences. Candidate A contains the QFr variant for 602 pairs and B for 603 (seed 42); every pair is evaluated again with the positions reversed. All eight LLMs receive one sentence pair per request, with one completed judgment retained per pair and candidate order (768-token generation limit). Open-weight models use Hugging Face Transformers on CUDA with greedy decoding; commercial models use their APIs. Reasoning is disabled or minimized where supported (Qwen3: disabled; GPT-5.4: none; Gemini: low). Invalid responses were subject to additional attempts or conservative recovery. For Llama, 172 judgments (7.14\% of its outputs) were recovered from malformed responses containing an explicit preference label. Recovery consisted of extracting the expressed preference and, where necessary, normalizing malformed equivalence labels to \texttt{EQUIVALENT}. No preference was inferred from the expected regional association, and missing evidence spans were left empty rather than reconstructed. The released predictions retain recovery indicators for these judgments. The complete evaluation prompt and JSON output format are provided in Appendix~\ref{app:evaluation_prompt}.

\textbf{Fine-tuned BERT classifiers.} Each encoder independently assigns an uncalibrated class-1 softmax score $s_m(x)\in[0,1]$ to each sentence, where higher scores indicate a stronger association with QFr. For model $m$ and sentence pair $i$, we define the score margin as $\Delta_{m,i}=s_m(x_i^{\mathrm{QFr}})-s_m(x_i^{\mathrm{MFr}})$. A positive margin indicates a correct ranking of the two variants, a negative margin indicates an incorrect ranking, and a zero margin indicates a tie. Sentence-level classification uses a threshold of 0.5, whereas contrastive ranking uses the sign of the score margin. Because each sentence is scored independently, candidate order does not affect the ranking.

\subsection{Evaluation Metrics and Statistical Analysis}

\textbf{Recognition accuracy.} For LLMs, we collapse strong and slight preference for \texttt{A} or \texttt{B} and count \texttt{EQUIVALENT} as incorrect under the benchmark's operational criterion. \textit{Mean single-order accuracy} averages the proportion of correct answers across both presentation orders, whereas \textit{order-robust accuracy} measures the proportion of sentence pairs with correct answers in both presentation orders. For encoders, \textit{strict pairwise accuracy} measures the proportion of pairs with $\Delta_{m,i}>0$, with ties counted as incorrect. For LLMs, independent uniform guessing between A and B yields expected single-order and order-robust accuracies of 50\% and 25\%, respectively. Uniform guessing among the five preference labels yields corresponding accuracies of 40\% and 16\%. For encoders, random ranking yields an expected strict pairwise accuracy of 50\%, assuming no ties. These baselines do not make the protocols equivalent. Across QFrBLiMa, the eight LLMs produce 19,280 judgments and the five encoders produce 12,050 scores and 6,025 pair rankings.

\textbf{Preference strength.} We distinguish five types of classifications: strong correct preference, slight correct preference, no preference, slight incorrect preference, and strong incorrect preference. For LLMs, these map directly to the returned preference labels and candidate correctness, using both presentation orders (2,410 judgments per model). For encoders, we partition the score margin $\Delta_{m,i}$ using symmetric thresholds: strong correct ($\Delta_{m,i}\geq0.20$), slight correct ($0.05\leq\Delta_{m,i}<0.20$), no preference ($|\Delta_{m,i}|<0.05$), slight incorrect ($-0.20<\Delta_{m,i}\leq-0.05$), and strong incorrect ($\Delta_{m,i}\leq-0.20$). These values were selected following sensitivity testing across 24 valid combinations of no-preference thresholds $\{0.01,0.025,0.05,0.075,0.10\}$ and strong-preference thresholds $\{0.10,0.15,0.20,0.25,0.30\}$, requiring the former to be strictly smaller than the latter. These thresholds provide a descriptive partition of score margins rather than empirically calibrated decision boundaries. Explicit LLM equivalence and limited encoder score separation are distinct outcomes; neither preference strength nor score margin constitutes calibrated confidence. The primary accuracy metrics remain unchanged. The complete threshold sensitivity results and encoder margin distributions are available in the accompanying repository.

\textbf{Recognition profiles.} Each linguistic marker is used in five contexts. Full, partial and no recognition denote correct classification of all five, one to four, or none of these contexts, respectively, using order-robust correctness for LLMs and strict pairwise ranking for encoders. For each thematic category, we calculate mean single-order and order-robust accuracy for each LLM and strict pairwise accuracy for each encoder, weighting all sentence pairs within the category equally. For each metric, we then compute the arithmetic mean across the corresponding group of models (eight LLMs or five encoders). Complementary coverage counts sentence pairs successfully recognized by at least one model within each group.

\textbf{Evidence spans.} An evidence span is a text excerpt returned by the LLM to support its decision. The evaluation prompt asks the LLM to return two evidence spans, one associated with QFr and the other with MFr. We evaluate whether these excerpts identify the documented variants of the linguistic marker and whether their regional associations are correct. We compute three proportions: \textit{grounding}, the number of judgments in which both returned spans occur in distinct candidate sentences; \textit{variant identification}, the number of judgments in which the returned spans match the two documented variants of the linguistic marker, regardless of their assigned regional labels; and \textit{span attribution}, the number of judgments in which both variants are identified and correctly assigned to their respective regional varieties. Each count is divided by the total number of judgments (2,410 per model). We also compute \textit{conditional decision accuracy} by dividing the number of correct final decisions among judgments with successful variant identification by the total number of judgments with successful variant identification. Unicode Normalization Form Compatibility Composition (NFKC), case folding and punctuation-insensitive tokenization support conservative matching. Missing spans, including equivalence responses, count as unsuccessful identification. Automatic matching does not constitute manual semantic validation.

\textbf{Statistical analysis.} We estimate 95\% percentile confidence intervals for the three primary accuracy metrics by resampling 241 linguistic-marker clusters with replacement (20,000 replicates; seed 42), retaining all five contexts and both candidate orders for LLMs. Paired marker-cluster sign-flip permutation tests (20,000 permutations) compare mean marker-level accuracy differences under exchangeability, with Holm correction separately across 28 LLM order-robust and ten encoder pairwise comparisons \citep{holm1979}. Exact paired McNemar tests compare original and inverted marginal accuracy, Holm-corrected across eight LLMs \citep{mcnemar1947}; these supplementary tests do not account for within-marker dependence. Cross-strategy comparisons and preference-strength analyses are descriptive.

\FloatBarrier
\section{Results and Discussion}

\subsection{Overall Recognition Performance}
\label{sec:model_performance}

\begin{figure}[!htbp]
    \centering
    \includegraphics[width=0.98\textwidth]{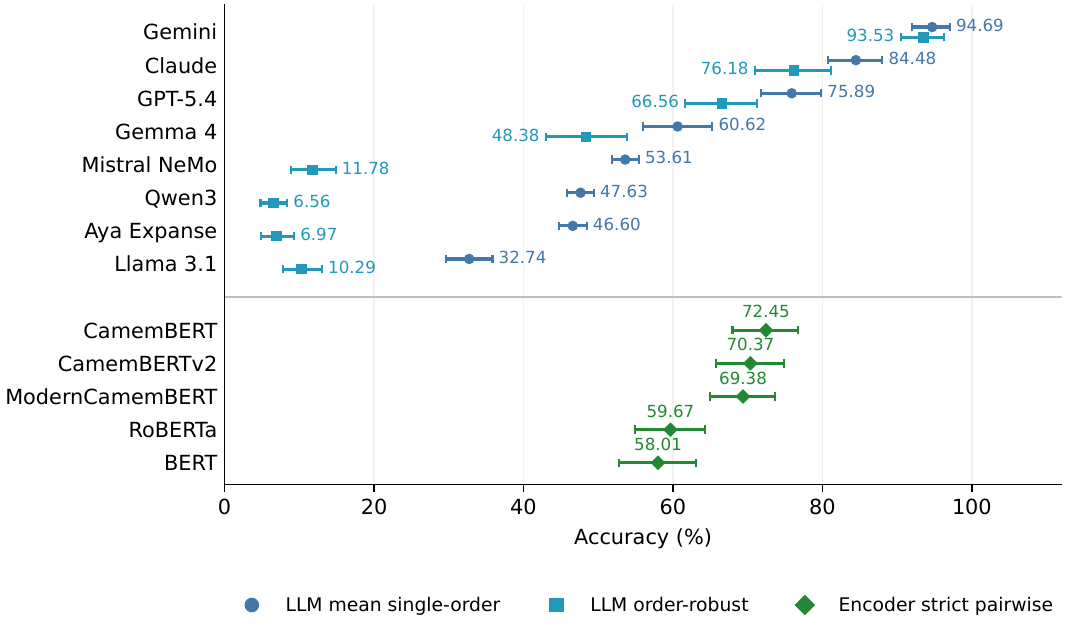}
    \caption{Overall recognition accuracy. LLM mean single-order and order-robust accuracy and encoder strict pairwise accuracy, with 95\% linguistic-marker-cluster bootstrap intervals. The evaluation protocols are distinct.}
    \label{fig:cross_family}
\end{figure}

We first compare the overall recognition accuracy of zero-shot LLMs and fine-tuned encoders on QFrBLiMa using the three accuracy metrics defined above. We then examine complementary coverage to determine whether the two strategies recognize different sentence pairs.

Gemini and Claude achieve higher order-robust accuracy than the strict pairwise accuracy of every encoder, and all three commercial LLMs exceed every encoder in mean single-order accuracy (Figure~\ref{fig:cross_family}). Conversely, the strict pairwise accuracy of each French-pretrained encoder exceeds the mean single-order accuracy of every open-weight LLM. Gemini leads the LLMs (93.53\% order-robust accuracy), while CamemBERT leads the encoders (72.45\% strict pairwise accuracy). Joint decisions and independent rankings remain methodologically distinct.

To determine whether differences in order-robust accuracy between LLMs are statistically significant, we perform paired linguistic-marker-cluster permutation tests across all 28 model comparisons. Of these, 23 remain significant after Holm correction. Gemini exceeds Claude by 17.34 pp, and Claude exceeds GPT-5.4 by 9.63 pp (both $p_{\mathrm{Holm}}=0.0014$); Mistral NeMo versus Aya Expanse does not reach significance ($p_{\mathrm{Holm}}=0.0512$). Gemini's robust 95\% interval (90.46--96.27\%) remains well above Gemma 4's (42.99--53.86\%) under linguistic-marker resampling (Figure~\ref{fig:cross_family}). These results characterize heterogeneity within the prompted strategy rather than statistical superiority over encoders.

To determine whether the two strategies recognize the same linguistic distinctions, we identify the sentence pairs correctly recognized by at least one model in each group. Of the 1,205 sentence pairs, 1,122 are recognized by both groups, 56 only by LLMs, 23 only by encoders and four by neither group. Encoders thus recover 23 of 27 pairs missed by all LLMs under their robust criterion. This complementarity shows that supervised classifiers retain discriminative capabilities not consistently observed in zero-shot judgments, even when their aggregate accuracy is lower. These counts depend on the number and selection of models within each group, as well as their respective success criteria.

\subsection{Preference Strength}
\label{sec:preference_strength}

To examine what aggregate accuracy conceals about decision strength, we compare the five preference categories defined in the experimental setup, using both candidate orders for LLMs and score-margin thresholds of 0.05 and 0.20 for encoders.

\begin{figure}[!htbp]
    \centering
    \includegraphics[width=0.98\textwidth]{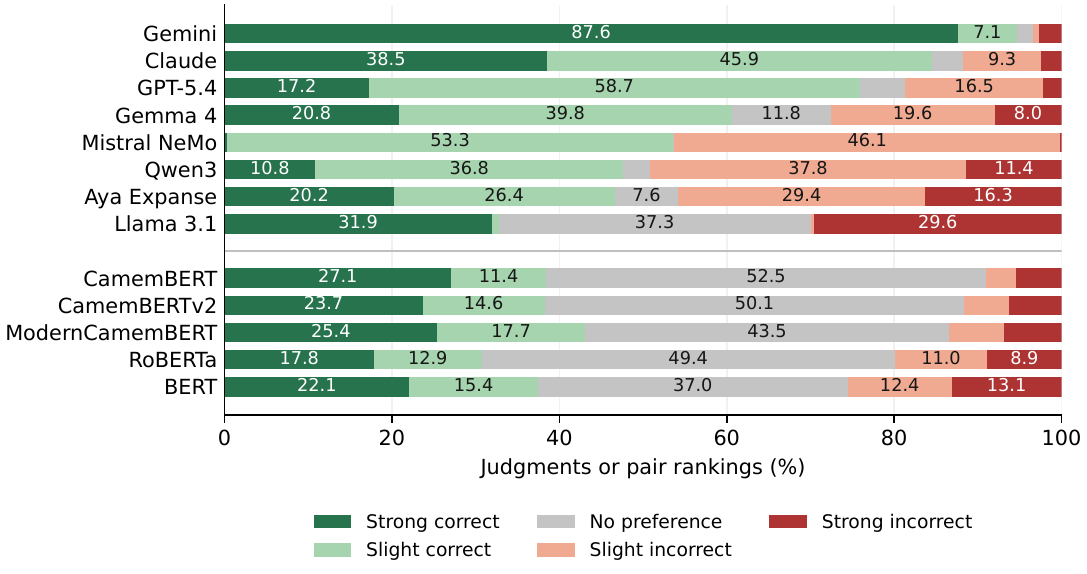}
    \caption{Distribution of five preference categories over 2,410 judgments per LLM and 1,205 pair rankings per encoder. Encoder categories use absolute score-margin thresholds of 0.05 and 0.20. No preference denotes explicit equivalence for LLMs and a small score margin for encoders.}
    \label{fig:preference_strength}
\end{figure}

As shown in Figure~\ref{fig:preference_strength}, Gemini expresses a strong correct preference in 87.59\% of judgments and a strong incorrect preference in only 2.70\%; its strong preferences are correct more often than its slight preferences (97.0\% versus 90.5\%). Claude and GPT-5.4 more frequently express slight correct preferences (45.93\% and 58.67\%) than strong correct preferences (38.55\% and 17.22\%). Mistral NeMo expresses slight preferences in 99.42\% of judgments, irrespective of correctness. Llama exhibits a contrasting profile: 37.30\% equivalence responses, 31.91\% strong correct preferences and 29.59\% strong incorrect preferences. Its low accuracy therefore reflects both frequent equivalence and strongly expressed incorrect attribution.

For encoders, the proportion of small margins ranges from 37.01\% (BERT-base-cased) to 52.53\% (CamemBERT). CamemBERT achieves the highest strict pairwise accuracy despite the smallest median absolute margin ($0.0427$), whereas BERT-base-cased achieves the lowest accuracy despite the largest median absolute margin ($0.0984$). Across the tested threshold combinations, BERT-base-cased consistently exhibits the highest proportion of strong incorrect rankings at a common strong-preference threshold, although absolute proportions depend on the thresholds. Thus, greater score separation does not necessarily indicate more accurate regional recognition. The preference categories describe decision strength rather than calibrated confidence, and small positive margins still count as correct under strict pairwise accuracy.

\subsection{Candidate-Order Sensitivity}
\label{sec:response_distribution}

To assess whether LLM decisions remain stable when candidate order changes, we compare the original and inverted judgments for every sentence pair, considering both marginal accuracy and paired correctness. We also examine candidate selection frequencies to distinguish stable recognition from systematic position preferences. A predictor always selecting A provides a reference for fixed-position behavior: approximately 50\% accuracy in each presentation order, exactly one correct judgment for every sentence pair, and zero order-robust accuracy.

\begin{figure}[!htbp]
    \centering
    \includegraphics[width=0.98\textwidth]{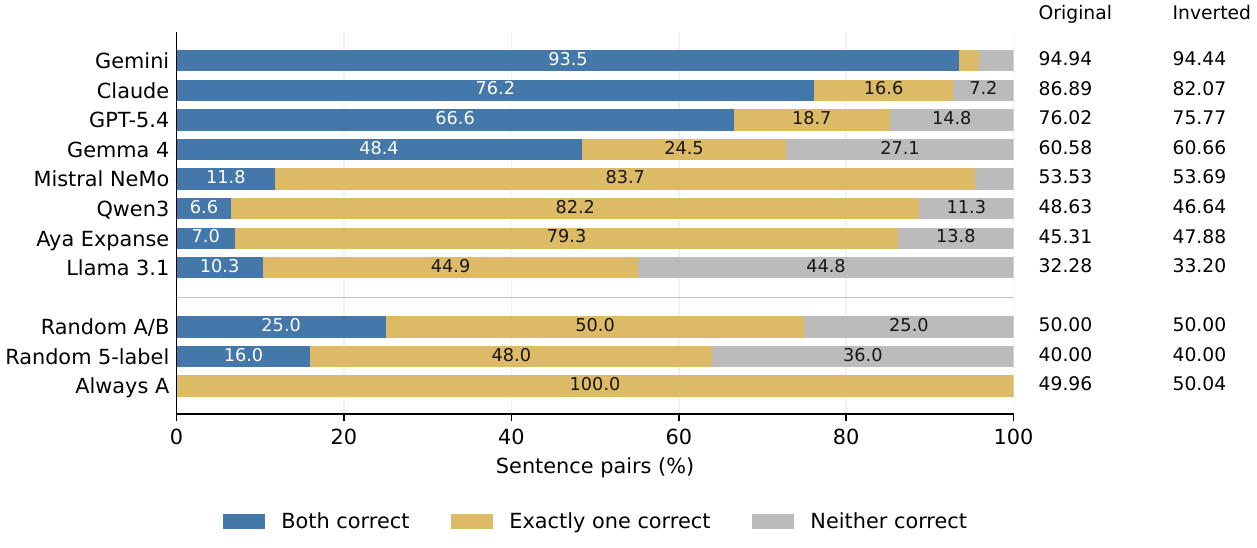}
    \caption{LLM outcomes across both candidate orders (\%): correct in both, exactly one, or neither presentation. Original and inverted marginal accuracies are shown alongside the distributions. Random A/B and Random 5-label assume independent uniform draws; Always A selects A in both orders.}
    \label{fig:candidate_order}
\end{figure}

In Figure~\ref{fig:candidate_order}, Mistral NeMo, Qwen3 and Aya Expanse succeed in exactly one order on 79.25--83.65\% of pairs, while their order-robust accuracy falls to 6.56--11.78\%. This discrepancy indicates that successful recognition under a single presentation order frequently fails to persist when the same linguistic distinction is presented in reverse. Their order-robust accuracies fall below both independent-guessing references (25\% and 16\%), suggesting that the observed instability is not adequately characterized by independent random errors. Indeed, despite balanced answer positions, the three models select A in 92.0/90.5\%, 87.3/86.3\% and 82.2/82.2\% of original/inverted judgments, respectively. The persistence of these selection preferences across both orders suggests that candidate placement or answer-label preferences can override the regional distinction being evaluated. This is consistent with previously reported position and answer-label biases \citep{wang-etal-2024-large-language-models-fair,pezeshkpour-hruschka-2024-large}, although our protocol does not isolate their respective contributions.

Llama exhibits a different recognition profile: it selects A in only 15.1/14.8\% of original/inverted judgments and fails in both orders on 44.81\% of pairs. Whereas the predominantly A-selecting models frequently recognize the expected regional association in one order but not the other, Llama more frequently fails to recognize it under either presentation. Low order-robust accuracy can therefore reflect distinct failure patterns, including systematic candidate preferences and persistent recognition failures. Moreover, no sentence pair is robustly recognized by all eight LLMs, indicating that recognition consistency varies across models even when the linguistic distinction remains unchanged. These findings highlight the importance of examining paired outcomes rather than relying exclusively on aggregate accuracy to characterize regional linguistic recognition.

Under the supplementary item-level McNemar analysis, only Claude exhibits a significant marginal accuracy difference after Holm correction (86.89\% versus 82.07\%, $p_{\mathrm{Holm}}=0.00040$). The three predominantly A-selecting models show no significant marginal difference despite substantial pair-level instability: losses under reversal are offset by gains on other items. Similar marginal accuracy therefore does not imply robust attribution. The robust criterion reveals pair-level instability that is not reflected in differences between marginal accuracies (Figure~\ref{fig:candidate_order}).

\subsection{Contextual Consistency}
\label{sec:contextual_consistency}

To assess whether regional recognition is consistent across different sentence realizations of the same linguistic marker, we count the linguistic markers recognized in all five, one to four, or none of their contexts under the respective success criterion of each strategy.

\begin{figure}[!htbp]
    \centering
    \includegraphics[width=0.98\textwidth]{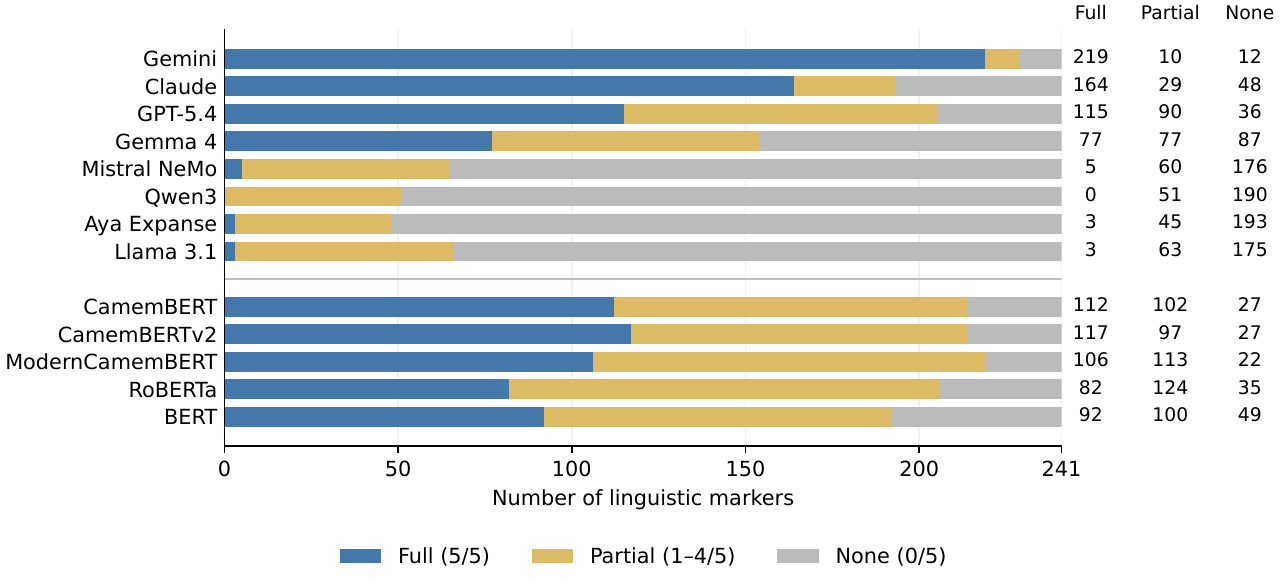}
    \caption{Number of linguistic markers recognized in all five, one to four, or none of their contexts. Success is defined by order-robust correctness for LLMs and strict pairwise ranking for encoders. Each model is evaluated on 241 linguistic markers.}
    \label{fig:contextual_consistency}
\end{figure}

Full recognition across five contexts (Figure~\ref{fig:contextual_consistency}) distinguishes consistent attribution from occasional success. Claude exhibits a more polarized recognition profile than GPT-5.4, fully recognizing 164 linguistic markers while failing in every context for 48, whereas GPT-5.4 shows more partial recognition. Among encoders, CamemBERTv2 fully recognizes more linguistic markers than CamemBERT (117 versus 112) despite lower overall pairwise accuracy, while ModernCamemBERT misses the fewest entirely (22). All encoders fully recognize numerically more linguistic markers than any open-weight LLM under their respective criteria. These profiles illustrate that aggregate accuracy, contextual consistency and lexical coverage capture distinct aspects of regional recognition: higher overall accuracy does not necessarily imply consistent recognition of more linguistic markers or broader lexical coverage. Aggregate rankings may therefore obscure complementary linguistic capabilities.

\subsection{Classification Transfer and French Pretraining}
\label{sec:encoder_results}

We assess whether source-based regional classification transfers to recognition of individual linguistic markers by comparing sentence-level macro-F1 on the internal test set and QFrBLiMa, using the same classification threshold of 0.5. We also examine external accuracy and contrastive ranking. To investigate the association between French pretraining and regional recognition, we compare the strict pairwise accuracy of French- and English-pretrained encoders under identical downstream supervision.

\begin{table}[!htbp]
\centering\small
\setlength{\tabcolsep}{6pt}
\begin{tabular}{lrrr}
\toprule
Model & Internal macro-F1 & External macro-F1 & External accuracy \\
\midrule
CamemBERT & \textbf{97.56} & \textbf{59.20} & \textbf{59.38} \\
CamemBERTv2 & 97.07 & 54.89 & 57.39 \\
ModernCamemBERT & 97.09 & 53.24 & 56.97 \\
RoBERTa-base & 96.36 & 53.68 & 54.27 \\
BERT-base-cased & 95.45 & 52.44 & 53.78 \\
\bottomrule
\end{tabular}
\caption{Internal test and external QFrBLiMa sentence-level macro-F1 and external sentence-level accuracy for the five encoders (\%).}
\label{tab:classification_transfer}
\end{table}

\textbf{Transfer to regional lexical recognition.} Internal macro-F1 exceeds 95\% for every encoder, yet external sentence-level accuracy and macro-F1 remain below 60\% (Table~\ref{tab:classification_transfer}). Source-level supervision may reward topic, register or provenance cues that do not distinguish two otherwise comparable sentences. Although sentence-level accuracy and strict pairwise accuracy measure different outcomes, the higher observed pairwise scores indicate that the encoders can often rank the QFr variant above its MFr counterpart even when their absolute classification decisions are less reliable. This internal-to-external performance gap complements previous work on French dialect identification \citep{bernier-colborne-etal-2022-transfer,bernier-colborne-etal-2023-dialect}. Whereas source-level classification measures whether a model can distinguish texts associated with different varieties, our contrastive evaluation shows that this ability does not necessarily imply reliable attribution of individual regional expressions.

\textbf{French-specific encoder pretraining.} All six French-versus-English encoder comparisons (Figure~\ref{fig:cross_family}) are significant after Holm correction ($p_{\mathrm{Holm}}\leq0.00225$; differences 9.71--14.44 percentage points). None of the three French-pretrained comparisons or the English-pretrained comparison is significant. The original CamemBERT has the highest observed pairwise accuracy, but its advantage over CamemBERTv2 (2.07 pp; $p_{\mathrm{Holm}}=1.0000$) or ModernCamemBERT (3.07 pp; $p_{\mathrm{Holm}}=0.7402$) does not establish superiority or equivalence. These findings suggest a potential benefit of French pretraining under identical downstream supervision, but do not establish a causal relationship. The French encoder ranking does not follow reported pretraining corpus size (Table~\ref{tab:pretraining}). Corpus composition, architecture, tokenization and optimization may all contribute.

\textbf{Multilingual LLM pretraining.} The limited robust performance of several multilingual LLMs is noteworthy: Llama supports French, Qwen3 covers 119 languages and dialects, and Aya Expanse targets 23 languages including French \citep{grattafiori2024llama,yang2025qwen3,dang2024aya}. Mistral NeMo and Gemma 4 also report multilingual capabilities (see the model cards linked in Section~\ref{sec:encoders_method}). General multilingual capability is therefore not sufficient for reliable regional attribution in this setting. The corpus-composition effects reported by \citet{khan2026quebec} motivate examining regional training exposure rather than overall scale alone. However, missing French-specific LLM token counts and QFr-specific counts for both families prevent a quantitative exposure comparison. Additional supervised examples and total multilingual tokens cannot explain the observed differences by themselves; model capacity and post-training also differ.

\subsection{Thematic Performance and Error Analysis}
\label{sec:marker_analysis}

To investigate whether recognition varies across lexical domains, we compute model-level accuracy and group-level means within the 12 thematic categories. We then examine individual linguistic markers with complementary or shared failures to characterize the linguistic distinctions associated with recognition difficulty.
For individual linguistic markers, pooled results comprise 40 model-context combinations for LLMs (eight models and five contexts) and 25 for encoders (five models and five contexts).

\begin{figure}[!htbp]
    \centering
    \includegraphics[width=1.0\textwidth]{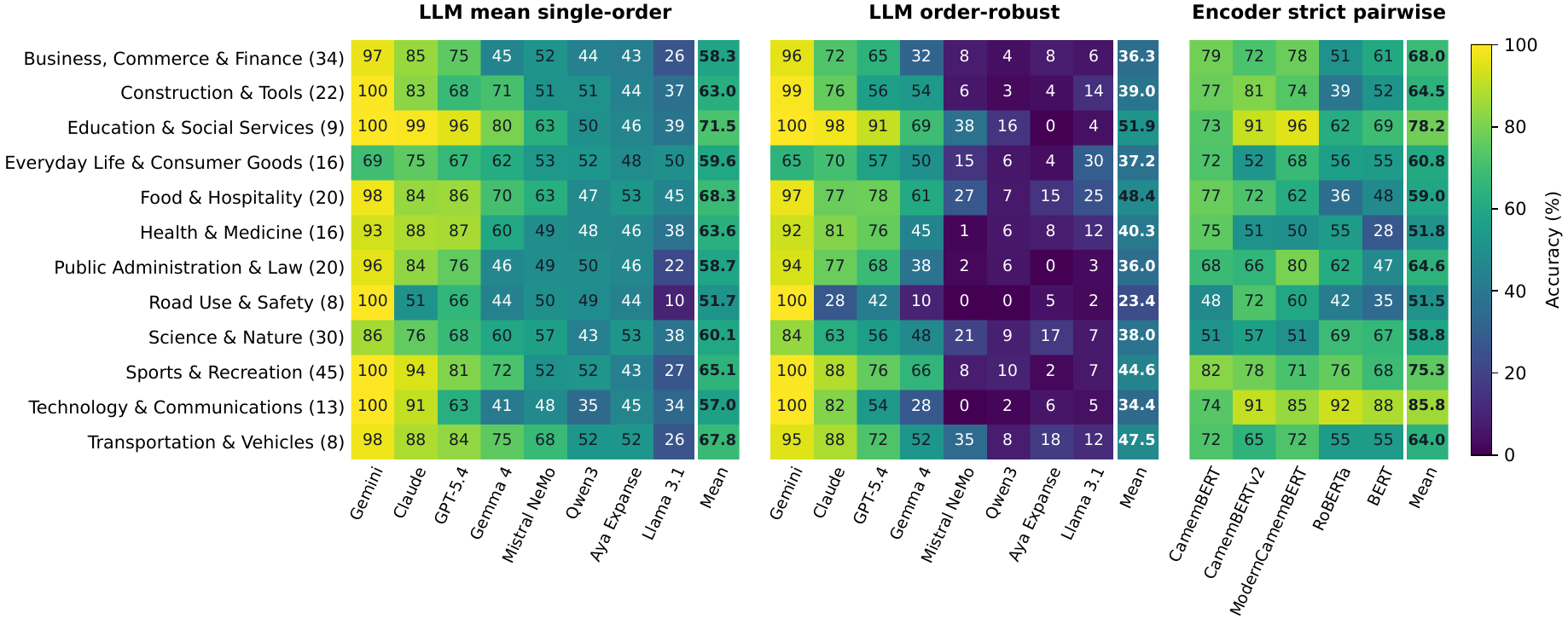}
    \caption{Thematic recognition profiles across 12 categories. Panels report LLM mean single-order accuracy, LLM order-robust accuracy and encoder strict pairwise accuracy (\%). Category labels indicate the number of linguistic markers. Individual model values are rounded to whole percentages; the Mean columns report group averages calculated before rounding, to one decimal place.}
    \label{fig:joint_domain_heatmap}
\end{figure}

The thematic profiles (Figure~\ref{fig:joint_domain_heatmap}) suggest that recognition performance varies substantially across models, while thematic categories reveal additional differences in their relative strengths and weaknesses. Technology and Communications is the highest-performing category for encoders (85.8\% strict pairwise accuracy), despite comparatively weak LLM performance. Conversely, Food and Hospitality is relatively difficult for encoders (59.0\% strict pairwise accuracy) but comparatively favorable to LLMs (48.4\% order-robust accuracy) relative to their performance across other categories. These observations motivate investigating domain-specific training exposure, although neither regional LLM pretraining composition nor linguistic-marker-level distribution within the supervised corpus has been established.

Category averages also conceal linguistic-marker-specific difficulties. Education and Social Services leads the LLM means, partly reflecting successful recognition of related expressions involving \textit{rel\^ache} (e.g., \textit{vacances universitaires/rel\^ache}, 27/40 robust LLM outcomes). Road Use and Safety has the lowest group mean under each of the three accuracy metrics, yet Gemini recognizes every sentence pair under both candidate orders, whereas Claude reaches 27.5\%, largely because it misses five linguistic markers involving \textit{att\'enuateur de choc/att\'enuateur d'impact} variants. In Everyday Life, Gemini's failures concentrate on \textit{d\'egrafeuse} and \textit{\'epingle \`a cravate}. These observations suggest that category-level performance may reflect consistent success or failure on a small set of linguistic markers, particularly when several contain lexically related variants. Since each linguistic marker contributes five sentence pairs, difficulties with related variants can disproportionately affect accuracy in categories containing few linguistic markers, without necessarily indicating limited knowledge of the entire domain.

\textbf{Complementary linguistic errors.} The mean single-order, robust and encoder pairwise accuracies for \textit{menu \`a prix fixe/table d'h\^ote} are 80\%, 60\%, and 20\%; for \textit{fixe-cravate/\allowbreak\'epingle \`a cravate}, the pattern reverses (32.50\%, 7.50\%, 80\%). The encoders rank \textit{b\^aton de hockey} above \textit{crosse} in every context but never rank \textit{b\^aton} above \textit{crosse de hockey}; LLM robust accuracy is 22.50\% for both. These examples suggest sensitivity to precise lexical realization, without identifying the classifiers' learned features. Percentages average eight LLMs or five encoders, with five contexts per linguistic marker.

To characterize persistent difficulties, we select the five linguistic markers with the highest pooled frequency of LLM failures in both orders and examine encoder rankings on those same linguistic markers.

\begin{table}[!htbp]
    \centering\small
    \begin{tabular}{lrrrr}
        \toprule
        & \multicolumn{2}{c}{Zero-shot LLM} & \multicolumn{2}{c}{Fine-tuned classifier} \\
        \cmidrule(lr){2-3}\cmidrule(lr){4-5}
        MFr / QFr variants & 0/2 correct & Rate & Incorrect ranking & Rate \\
        \midrule
        \textit{airelle ponctu\'ee / pomme de terre} & 23/40 & 57.5\% & 17/25 & 68.0\% \\
        \textit{serre-file / chercheur} & 21/40 & 52.5\% & 20/25 & 80.0\% \\
        \textit{petit-duc indien / petit-duc des Indes} & 19/40 & 47.5\% & 25/25 & 100.0\% \\
        \textit{cl\'e plate / cl\'e ouverte} & 19/40 & 47.5\% & 18/25 & 72.0\% \\
        \textit{crosse de hockey / b\^aton} & 18/40 & 45.0\% & 25/25 & 100.0\% \\
        \bottomrule
    \end{tabular}
    \caption{Encoder failures on the five linguistic markers selected for high pooled persistent LLM failure. The LLM 0/2 outcome includes incorrect selections and equivalence responses, whereas encoder failures include reversed score orderings and ties. The denominators and failure criteria differ.}
    \label{tab:difficult}
\end{table}

Shared failures (Table~\ref{tab:difficult}) illustrate distinct challenges for regional lexical recognition. In \textit{airelle ponctu\'ee/pomme de terre}, the QFr term denotes an uncommon botanical sense rather than the familiar potato; similarly, \textit{chercheur} in \textit{serre-file/chercheur} denotes a fire-evacuation role rather than a researcher. Correct regional attribution may therefore depend on resolving uncommon senses of familiar words. Subtle differences in \textit{petit-duc indien/petit-duc des Indes} and \textit{cl\'e plate/cl\'e ouverte} may instead require specialized terminological knowledge. However, specialized vocabulary is not uniformly difficult: \textit{perche truit\'ee/achigan \`a petite bouche} succeeds for 30/40 robust LLM outcomes and 24/25 encoder rankings. Together with the contrasting results for \textit{b\^aton de hockey} and \textit{b\^aton}, these examples suggest that familiarity with a concept does not necessarily imply knowledge of its regional lexical realizations. These retrospective observations provide exploratory explanations, not causal evidence, and complement the grammatical difficulties examined in QFrBLiMP \citep{beauchemin-etal-2026-qfrblimp}. Given the unequal number of linguistic markers across thematic categories, these comparisons are descriptive and should not be interpreted as statistically established domain-level differences.

\subsection{Evidence Span Identification and Attribution}
\label{sec:evidence_spans}

We investigate whether LLM errors arise when models fail to identify the relevant linguistic marker or when they identify its variants but fail to assign them to the correct regional varieties or select the correct candidate. To distinguish these outcomes, we evaluate the evidence spans returned by the LLMs using the four metrics defined in the experimental setup.

\begin{table}[!htbp]
    \centering
    \scriptsize\setlength{\tabcolsep}{2pt}
    \begin{tabular}{@{}p{0.28\columnwidth}rrrr@{}}
        \toprule
        \textbf{Model}
        & \textbf{Ground.}
        & \textbf{Ident.}
        & \textbf{Span attr.}
        & \textbf{Acc.$\mid$Ident.} \\
        \midrule
        Gemini 3.8 Flash
        & 97.93 & 85.77 & \textbf{82.32} & \textbf{95.98} \\
        Claude Opus 4.8
        & 96.10 & \textbf{89.54} & 78.51 & 87.63 \\
        GPT-5.4
        & 94.15 & 78.30 & 62.24 & 79.49 \\
        Gemma 4 12B IT
        & 87.76 & 69.96 & 49.13 & 70.23 \\
        Mistral NeMo 12B
        & \textbf{98.05} & 78.38 & 45.52 & 54.16 \\
        Qwen3 8B
        & 96.14 & 87.26 & 42.78 & 49.07 \\
        Aya Expanse 8B
        & 81.00 & 76.80 & 39.13 & 50.78 \\
        Llama 3.1 8B
        & 60.91 & 52.07 & 26.64 & 49.96 \\
        \bottomrule
    \end{tabular}
     \caption{Evidence span evaluation across eight LLMs
    (\%). Ground. denotes grounded evidence spans;
    Ident. denotes successful identification of both
    documented variants, independently of their assigned
    regional labels; Span attr. denotes correct regional
    attribution of both identified variants; and
    Acc.$\mid$Ident. denotes candidate-selection accuracy
    conditional on successful variant identification.
    The first three metrics are computed over all 2,410 judgments per model; the last is computed only over judgments with successful variant identification.}
    \label{tab:evidence_spans}
\end{table}

Qwen3 identifies both variants more frequently than Gemini (87.26\% versus 85.77\%), but its conditional decision accuracy remains near 50\% (49.07\%), compared with Gemini's 95.98\%. Mistral NeMo and Aya also identify both variants in over three quarters of judgments, with conditional accuracies near 50\%. Qwen3 and Mistral identify both variants in 84.87\% and 77.46\% of their incorrect judgments, respectively. Llama's lower identification rate (52.07\%) partly reflects frequent equivalence responses; among non-equivalence judgments it identifies both variants in 83.06\%. Thus, successfully identifying the documented variants does not guarantee correct regional attribution or candidate selection (Table~\ref{tab:evidence_spans}). The 172 recovered Llama judgments also contain empty evidence spans, which are counted as unsuccessful identification under our evaluation protocol.

Correct evidence attribution can also coexist with an incorrect preference label (122 Aya and 94 Mistral judgments). For example, Mistral NeMo correctly identifies \textit{casse-t\^ete} (B) as QFr and \textit{puzzle} (A) as MFr, yet selects \texttt{A\_SLIGHTLY\_MORE\_FQ}. Other rationales for \textit{fixe-cravate/\allowbreak\'epingle \`a cravate} identify both variants but reverse their geographic associations. These distinct errors separate variant identification, regional association and consistent final decisions; their underlying mechanisms remain untested.

\section{Conclusion}
QFrBLiMa provides a controlled evaluation of documented regional associations of linguistic markers through 1,205 contrastive sentence pairs. Across 13 models, zero-shot prompting and supervised fine-tuning exhibit complementary recognition patterns under their respective evaluation protocols. Although the best-performing commercial LLMs achieve the highest observed recognition accuracies, supervised classifiers remain competitive and successfully recognize some sentence pairs missed by all evaluated LLMs under the order-robust criterion. Our analyses distinguish several aspects of regional linguistic recognition that aggregate accuracy alone cannot capture: preference strength, robustness to candidate order, consistency across contexts, and correct identification and attribution of regional variants. Source-based supervision transfers imperfectly to contrastive recognition, while the stronger performance of French-pretrained encoders suggests an association between pretraining characteristics and regional recognition without establishing causality. Together, these findings demonstrate the value of dedicated contrastive benchmarks for characterizing the strengths and limitations of language models beyond general multilingual capabilities and source-based classification performance. The QFrBLiMa methodology can be adapted to other linguistic varieties for which suitable documented resources exist, particularly those underrepresented in NLP research.

\section*{Limitations}
Broader Canadian and European geographic labels may obscure finer regional distinctions. The GDT reflects institutional terminology rather than the full distribution of natural QFr. Documented regional associations do not imply exclusive usage, contemporary frequencies or natural production in unconstrained generation. Constructed contexts were not independently validated, and exposure to the GDT during pretraining cannot be excluded.

Order robustness conflates presentation effects, label preferences and inference variability, without repeated-run evaluation. LLM preference labels and encoder score margins are not directly comparable or calibrated confidence measures. Automatic evidence-span matching does not establish rationale faithfulness, and the recovery of malformed Llama responses limits evidence-span completeness.

Source-based supervision may capture topic, genre or provenance rather than regional linguistic characteristics. Document grouping is incomplete, and exact corpus reconstruction remains unverified. Model selection, a single training seed and incomplete pretraining disclosures limit generalizability. Statistical uncertainty is conditional on the evaluated configurations, and clustering by linguistic marker does not account for dependencies across markers from the same GDT entry. Because architecture, scale, pretraining and supervision are not independently controlled, cross-strategy comparisons remain descriptive rather than causal.

\section*{Ethical Considerations}
Regional labels do not imply linguistic superiority, exclusive usage or speaker identity. No personal participant data were collected for the benchmark. Existing web and other training sources have source-specific access and usage conditions documented in the repository. The GDT datasets are distributed under the Creative Commons Attribution-NonCommercial-ShareAlike 4.0 International license (CC BY-NC-SA 4.0) \citep{oqlf_data}. As an adaptation of these data, QFrBLiMa retains OQLF attribution and is distributed under the same license.

\section*{Acknowledgments}
This research was supported by the Natural Sciences and Engineering Research Council of Canada (NSERC). The authors thank the Coopérative nationale de l'information indépendante (CN2i), Le Soleil, and Radio-Canada for granting access to the resources used to construct the supervised corpus. The authors also thank the Office québécois de la langue française (OQLF) for making the terminological resources underlying QFrBLiMa publicly available.

\begin{aidisclosure}
OpenAI GPT-6.2 and Anthropic Claude Sonnet 5.5 were used to proofread the manuscript, improve linguistic clarity, and assist with LaTeX formatting and plotting scripts. The authors reviewed and verified all AI-assisted revisions and code and retain full responsibility for the methodology, analysis, reported results, and scientific conclusions.
\end{aidisclosure}

\printbibliography[heading=subbibintoc]

\newpage
\appendix
\section{Evaluation Prompt and Output Format}
\label{app:evaluation_prompt}

This appendix reproduces the system prompt and user prompt template used in the zero-shot LLM evaluation, as defined in \texttt{qfrblima/protocol.py} in the accompanying repository. All eight LLMs were evaluated using one sentence pair per request, independently in both candidate orders. The original terminology is retained: International French (FI) and Quebec French (FQ) correspond to MFr and QFr, respectively, in the benchmark protocol. Placeholder values illustrate the input serialization.

\textbf{System prompt.}
\begin{Verbatim}[fontsize=\footnotesize,breaklines=true,breakanywhere=true]
You are acting as an expert linguistic evaluator of International French (FI)
and Quebec French (FQ).

For every item, you receive exactly two anonymized sentences: Candidate A and
Candidate B. They express the same underlying content with a small linguistic
contrast.

PRIMARY TASK
Determine which candidate, if either, contains MORE LINGUISTIC CHARACTERISTICS
COMPATIBLE WITH QUEBEC FRENCH.

IMPORTANT
- Candidate order is randomized.
- You are not told which candidate is FQ or FI.
- Do not infer the answer from position.
- Do not assume Quebec French means slang or joual.
- Quebec French may be colloquial, standard, institutional, technical, or formal.
- Small but genuine lexical, phraseological, syntactic, or register differences
  are meaningful.
- Do not invent a difference when there is no clear evidence.

DIMENSIONS
1. lexical:
   vocabulary or terminology characteristic of Quebec French usage.
2. phraseological:
   multi-word expressions, idioms, collocations, or conventional turns of phrase.
3. syntactic_register:
   constructions or register choices compatible with Quebec French.
4. mixed:
   more than one dimension clearly contributes.
5. none:
   no meaningful FQ-oriented difference can be identified.

OVERALL PREFERENCE
Choose exactly one:
- A_MUCH_MORE_FQ
- A_SLIGHTLY_MORE_FQ
- EQUIVALENT
- B_SLIGHTLY_MORE_FQ
- B_MUCH_MORE_FQ

Use MUCH when the difference is strong and unambiguous.
Use SLIGHTLY when the difference is genuine but limited.
Use EQUIVALENT only when neither candidate shows a meaningful FQ advantage.

EVIDENCE SPANS
- fq_evidence_span: copy the shortest exact span from the candidate you consider
  more compatible with Quebec French that supports your decision.
- fi_evidence_span: when possible, copy the corresponding shortest exact span
  from the other candidate.
- If overall_preference is EQUIVALENT, return an empty string for both spans.
- Do not explain inside the evidence fields; copy text only.

RATIONALE
Give one concise linguistic explanation for the decision.

Your goal is not to guess a dataset label. Your goal is to identify genuine
linguistic evidence of Quebec French usage from the two sentences alone.
\end{Verbatim}

\textbf{User prompt}
\begin{Verbatim}[fontsize=\footnotesize,breaklines=true,breakanywhere=true]
Evaluate every pair independently according to the protocol.

Do not assume that either candidate must be Quebec French. Use only linguistic evidence contained in the two sentences.

Return a JSON object with a "judgments" array containing exactly one entry. Use keys judgment_id, overall_preference, dominant_dimension, fq_evidence_span, fi_evidence_span, rationale. No Markdown.

<input_items>
[{"judgment_id": "<item_id>", "Candidate_A": "<sentence_A>", "Candidate_B": "<sentence_B>"}]
</input_items>
\end{Verbatim}

\textbf{JSON output example.} The response contains a \texttt{judgments} array with one entry per input item. Each entry has six required string fields; the preference and dimension values are restricted to the categories defined in the system prompt. The example below illustrates the structure rather than an observed model response.
\begin{Verbatim}[fontsize=\footnotesize,breaklines=true,breakanywhere=true]
{
  "judgments": [
    {
      "judgment_id": "<item_id>",
      "overall_preference": "A_SLIGHTLY_MORE_FQ",
      "dominant_dimension": "lexical",
      "fq_evidence_span": "<exact text span>",
      "fi_evidence_span": "<exact text span>",
      "rationale": "<short linguistic explanation>"
    }
  ]
}
\end{Verbatim}

\end{document}